\documentclass{article}

\usepackage{arxiv}

\usepackage[T1]{fontenc}    
\usepackage{hyperref}       
\usepackage{url}            
\usepackage{booktabs}       
\usepackage{amsfonts}       
\usepackage{nicefrac}       
\usepackage{microtype}      
\usepackage{lipsum}		
\usepackage{graphicx}
\usepackage{natbib}
\usepackage{doi}
\usepackage{graphicx}%
\usepackage{multirow}%
\usepackage{amsmath,amssymb,amsfonts}%
\usepackage{mathrsfs}%
\usepackage[title]{appendix}%
\usepackage{xcolor}%
\usepackage{textcomp}%
\usepackage{manyfoot}%
\usepackage{booktabs}%
\usepackage{algorithm}%
\usepackage{algorithmicx}%
\usepackage{algpseudocode}%
\usepackage{listings}%
\usepackage{amssymb}
\usepackage{subfigure}
\usepackage{courier}
\usepackage{amsmath}
\usepackage{array}
\usepackage[latin1]{inputenc}
\usepackage{multirow}
\usepackage{graphicx}
\usepackage{rotating}
\usepackage{amssymb}
\usepackage{amsthm}
\usepackage[T1]{fontenc}

\theoremstyle{thmstyleone}%
\theoremstyle{thmstyletwo}%

\theoremstyle{thmstylethree}%
\newtheorem{definition}{Definition}%

\title{Accelerating prototype selection with spatial abstraction}

\author{ \href{https://orcid.org/0000-0002-4499-3601}{\includegraphics[scale=0.06]{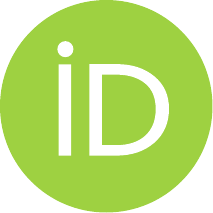}\hspace{1mm}Joel Lu{\'i}s Carbonera}\thanks{\url{https://scholar.google.com/citations?user=oMtHaoQAAAAJ&hl}} \\
	Institute of Informatics \\
	Federal University of Rio Grande do Sul (UFRGS)\\
	Porto Alegre, Brazil \\
	\texttt{jlcarbonera@inf.ufrgs.br} \\
}

\renewcommand{\shorttitle}{\textit{arXiv} Template}

\hypersetup{
pdftitle={A template for the arxiv style},
pdfsubject={q-bio.NC, q-bio.QM},
pdfauthor={David S.~Hippocampus, Elias D.~Striatum},
pdfkeywords={First keyword, Second keyword, More},
}

\begin{document}
\maketitle

\begin{abstract}
The increasing digitalization in industry and society leads to a growing abundance of data available to be processed and exploited. However, the high volume of data requires considerable computational resources for applying machine learning approaches. Prototype selection techniques have been applied to reduce the requirements of computational resources that are needed by these techniques. In this paper, we propose an approach for speeding up existing prototype selection techniques. It builds an abstract representation of the dataset, using the notion of spatial partition. The second step uses this abstract representation to prune the search space efficiently and select a set of candidate prototypes. After, some conventional prototype selection algorithms can be applied to the candidates selected by our approach. Our approach was integrated with five conventional prototype selection algorithms and tested on 14 widely recognized datasets used in classification tasks. The performance of the modified algorithms was compared to that of their original versions in terms of accuracy and reduction rate. The experimental results demonstrate that, overall, our proposed approach maintains accuracy while enhancing the reduction rate of the original prototype selection algorithms and simultaneously reducing their execution times.
\end{abstract}

\keywords{Prototype selection \and Instance selection \and Data reduction \and Machine learning}

\section{Introduction}\label{sec:introduction}
    The transition to digital information within industry and society has resulted in a growing abundance of accessible data. This vast volume of data poses challenges to data mining and machine learning techniques, as scalability becomes a significant concern in such contexts, as noted by \cite{fan2013mining}. Consequently, strategies for \emph{data reduction} have been employed to address the complexities posed by large datasets in this environment.
	
    \emph{Prototype selection} is as \emph{data reduction} technique that can applied in machine learning tasks. The main goal of prototype selection approaches is to generate a more concise set of representative data points from the entirety of available data. It is worth noting that as these approaches discard a portion of the available samples, they inherently encounter a \emph{trade-off} between reducing the dataset size (thus lowering computational resource requirements) and preserving the \emph{quality} of classification results, as outlined by \cite{CHOU:2006,carbonera2018efficient}. 
	
    Numerous studies have indicated that prototype selection not only yields manageable datasets, thereby reducing the computational resources required for machine learning tasks, but it may also enhance the accuracy of the resulting models. This improvement stems from the elimination of redundant and/or erroneous instances from the training set, as highlighted by \cite{de2019instance}. Besides that, recent works \cite{alswaitti2024training} highlight prototype selection algorithms as enablers of sustainable AI practices for training \emph{green AI models}.
	
    Some works, such as \cite{arnaiz2016instance}, distinguish \emph{prototype selection} from \emph{instance selection}. In this perspective, \emph{instance selection} selects a subset of the original instances in the dataset, whereas \emph{prototype selection} can replace the original instances by \emph{synthetic ones}. In this work, we adopt the expression \emph{prototype selection} to encompass both kinds of approaches. 
	
    Most of the approaches for prototype selection proposed in the literature, such as \cite{WILSON:1972,WILSON:2000,BRINGHTON:2002,NIKOLAIDIS:2011,LIN:2015,LEYVA:2015}, suffer from a \emph{high execution time}, posing challenges particularly when handling large datasets. In this paper, we propose a simple approach called PSASA (\emph{P}rototype \emph{S}election \emph{A}ccelerator based on \emph{S}patial \emph{A}bstraction) designed for speeding up conventional prototype selection algorithms. The approach can be viewed as a pre-step that produces in an \emph{efficient way} an initial set of candidate prototypes that is smaller than the original set. The algorithm has two main steps: (I) it adopts the notion of \emph{spatial partition} for identifying sets of instances that are similar to each other in the original dataset; after, (II) for each spatial partition, it creates a prototype from the set of instances of each class within that partition. Then, the set of prototypes produced by this process can be used as input for some conventional prototype selection algorithms in a subsequent step. Thus, our approach performs a rough (and fast) prototype selection that can be refined by another conventional (and, in general, slower) prototype selection algorithm. Moreover, since the input of the second algorithm (some conventional instance selection algorithm) is reduced, its running time is reduced proportionally. To our knowledge, this is the first proposal of a pre-step for accelerating conventional prototype selection approaches. 
	
    Notice that the main focus of our approach is to accelerate the processing time of conventional prototype selection algorithms, trying to keep its original performance regarding the classification quality achieved by a classifier trained with the resulting reduced dataset. Therefore, our objective is not to improve the classification quality of the traditional algorithms.
	
    The approach proposed in this paper is based on notions proposed in \cite{carbonera2018efficient,carbonera2018efficient1,carbonera2018efficient2}. However, it is important to notice that these previous works propose novel prototype selection approaches, whereas our paper presents a novel algorithm aimed at accelerating conventional prototype selection methods.
	
    We integrated our approach with five conventional prototype selection algorithms sourced from existing literature. Subsequently, we evaluated the resulting methodologies in a classification task using 14 widely recognized datasets. Their performances were then compared against the original versions of the five algorithms, with assessments conducted based on two distinct performance metrics: accuracy and reduction. The accuracy of the algorithms was assessed using SVM and KNN classifiers. The findings demonstrate that our approach accelerates the identified prototype selection algorithms, resulting in lower execution times, while yielding reduction rates and accuracies that are comparable, and in certain instances, superior to those achieved by the original algorithms.
	
    Section \ref{sec:relatedworks} presents some related works. Section \ref{sec:notations} presents the notation that will be used throughout the paper. Section \ref{sec:ourApproach} presents our approach. Section \ref{sec:experiments} discusses our experimental evaluation. Finally, Section \ref{sec:conclusion} presents our main conclusions and final remarks.

\section{Related works} \label{sec:relatedworks}
	
    Over the past decades, several different approaches for prototype selection have been proposed. The \emph{Condensed Nearest Neighbor} algorithm \cite{HART:1968} and \emph{Reduced Nearest Neighbor} algorithm (RNN) \cite{GATES:1972} are examples of some of the earliest proposals for prototype selection. However, both exhibit similar drawbacks: they may include noisy instances in the final selected set, their outcomes are sensitive to the sequence of instances, and they are characterized by high time complexity. 
    
    The \emph{Edited Nearest Neighbor} (ENN) algorithm \cite{WILSON:1972}, which is a widely applied approach, removes instances that do not agree with the label of the majority of its $k$ nearest neighbors. Since the main goal of this strategy is removing noisy instances, its reduction rate is low in comparison to the other algorithms. 
    
    In \cite{WILSON:2000}, the authors present the well-known DROP3 (\emph{Decremental Reduction Optimization Procedure}). In its first step, this algorithm applies a noise-filter algorithm such as ENN. Then, it removes an instance $x$ if the instances that have $x$ as one of its $k$ nearest neighbors in the original training set can be correctly classified without $x$. Despite its effectiveness and widespread application, DROP3 is characterized by a high time complexity.
    
    The \emph{Iterative Case Filtering algorithm} (ICF) is proposed in \cite{BRINGHTON:2002}. This algorithm assumes that the set of instances in $T$ whose distance from $x$ is less than the distance between $x$ and its nearest enemy (instance with a different class) is the \emph{coverage set} of $x$. On the other hand, the set of instances in $T$ that have $x$ in their respective coverage sets is the \emph{reachable set} of $x$. The algorithm removes an instance $x$ from $S$ when $|Reachable(x)| > |Coverage(x)|$. The main drawback of ICF is its high running time. 
    
    In \cite{LEYVA:2015}, a collection of techniques for prototype selection was introduced, all grounded in the concept of a \emph{local set}. The local set of a given instance $x$ is the set of instances contained in the largest hypersphere centered on $x$ that does not include any instance with a class that is different from the class of $x$. One of the algorithms, called \emph{Local Set-based Smoother} (LSSm) employs the notions of \emph{usefulness} and \emph{harmfulness} for guiding the process. In this context, the usefulness ($u(x)$) of a given instance $x$ is the number of instances having $x$ among the members of their local sets, and the harmfulness ($h(x)$) is the number of instances having $x$ as its \emph{nearest enemy}. The algorithm includes a given $x$ in $S$ if $u(x) \geq h(x)$. Notice that the reduction rate of LSSm is lower than most of the instance selection algorithms, since its main goal is just removing noisy instances from the dataset. In \cite{LEYVA:2015} the authors propose also another instance selection algorithm called \emph{Local Set Border selector} (LSBo). It firstly uses LSSm to remove noise and then, it computes the local set of every instance $\in$ $T$. In the last step, LSBo includes in $S$ every instance $x \in T$ whose local set does not include any instance already included in $S$. The time complexity of the two approaches is $O(|T|^2)$. In \cite{carbonera2019local}, the authors propose an improvement of LSBo called ASBI (Approach for Selection of Border Instances). This algorithm adopts the notions of \emph{local set} (as LSBo), \emph{internality}, \emph{degree of potential noise}, \emph{coherence} and \emph{seletivity} for selecting the most representative instances of each class. The authors show that ASBI surpasses the original LSBo algorithm in accuracy and reduction rate.

    The notion of \emph{density} is also employed by several different algorithms for guiding the prototype selection. The density of a given instance $x$ is related to the average distance of $x$ to all the instances in its neighborhood (that can be the whole dataset, in some cases). Some of the first algorithms that adopt this notion of density for prototype selection are LDIS (Local Density-based Instance selection) \cite{CARBONERA2015ICTAI}, CDIS (Central Density-based Instance Selection) \cite{carbonera2016novel}, XLDIS (extended local density-based instance selection) \cite{carbonera2017efficient} and DPS (Density-based Prototype Selection) \cite{carbonera2020density}, which preserves only the locally densest instances in each class. This strategy ensures that these algorithms achieve a high reduction rate and relatively low time complexity. However, the main drawback of both algorithms is that, in some cases, they can harm the classification quality. In order to overcome this issue, in \cite{malhat2020new} the authors propose the GDIS (global density-based instance selection) and EGDIS (enhanced global density-based instance selection). Both algorithms adopt a strategy that is similar to those adopted in LDIS and CDIS, but consider a global density measure to evaluate the best instances for representing the dataset. This perspective achieves better classification quality.

    In \cite{carbonera2020attraction}, the authors propose AIS (Attraction-based Instance selection). This algorithm is inspired by LDIS \cite{CARBONERA2015ICTAI} and adopts the notion of \emph{attraction} for selecting the most representative instances of each class. The main advantage of AIS in comparison with the original LDIS algorithm is the fact that AIS allows the user to define the percentage of instances of the original dataset that should be preserved.
    
    Some approaches \cite{babu2001comparison,chen2015instance}, on the other hand, apply some metaheuristics for selecting a set of prototypes that maximize the accuracy. In \cite{moran2022curious} the authors combine reinforcement learning and clustering techniques for prototype selection.

    Other approaches, such as \cite{carbonera2018efficient, carbonera2018efficient1, carbonera2018efficient2} are based on the notion of \emph{spatial abstraction}, which involves splitting the overall data space in a set of \emph{spatial partitions} and selecting prototypes in each partition independently. These notions inspired the approach proposed in this work for accelerating conventional prototype selection algorithms.
    
    A recent survey and experimental evaluation on different prototype selection approaches is provided by \cite{cunha2023comparative}.

\section{Notations}\label{sec:notations}

	In this section, we present the notation adapted from \cite{CARBONERA2015ICTAI} that will be used throughout the paper. 
	\begin{itemize}
		\item $T = \{o_\mathit{1},o_\mathit{2},...,o_\mathit{n}\}$ is a non-empty set of $n$ instances representing the original dataset that will be reduced in the prototype selection process.
		
		
		\item $D = \{d_\mathit{1},d_\mathit{2},...,d_\mathit{m}\}$ is a set of $m$ dimensions (that represent features or attributes), where each $d_\mathit{i} \subseteq \mathbb{R}$.
		
		\item Each $o_\mathit{i} \in T$ is an $m-tuple$, such that $o_\mathit{i} = (o_\mathit{i1}, o_\mathit{i2},...,o_\mathit{im})$, where $o_\mathit{ij}$ represents the value of the $j$-th dimension of the instance $o_\mathit{i}$, for $1 \leq j \leq m$.
		
		
		\item $L = \{l_\mathit{1},l_\mathit{2},...,l_\mathit{p}\}$ is the set of $p$ class labels that are used for classifying the instances in $T$, where each $l_\mathit{i} \in L$ represents a given class label. 
		
		
		
	\end{itemize}
	
\section{The proposed approach} \label{sec:ourApproach}

In this paper, we propose the PSASA (Prototype selection accelerator based on spatial abstraction) algorithm designed for \emph{speeding up} conventional algorithms of prototype selection. Our approach adopts an efficient strategy for producing a set of candidate prototypes, which is smaller than the original set of instances. In a subsequent step, some conventional prototype selection algorithms can be applied to refine this set, producing the final set of selected prototypes. Thus, our approach can be viewed as a pre-processing step whose role is to efficiently reduce the search space for the subsequent application of some conventional prototype selection approach.

Since our approach is based on the notion of \emph{spatial partition} \cite{carbonera2018efficient,carbonera2018efficient1,carbonera2018efficient2}, it is important to introduce it before discussing the proposed approach. 
\begin{definition}
A \emph{spatial partition} of a spatial region that contains a given set of objects $H \subseteq T$ is a set $\mathcal{SP_\mathit{H}} = \{s_\mathit{1},s_\mathit{2},...,s_\mathit{m}\}$, where:
\begin{equation}
\forall s_\mathit{i} \in \mathcal{SP_\mathit{H}} \rightarrow \exists d_\mathit{i} \in D \wedge s_\mathit{i} \subseteq d_\mathit{i}    
\end{equation}
\begin{equation}
\forall d_\mathit{i} \in D \rightarrow \exists s_\mathit{i} \in \mathcal{SP_\mathit{H}}
\end{equation}
\begin{equation}
\begin{split}
\forall s_\mathit{i} \in \mathcal{SP_\mathit{H}} \rightarrow & s_\mathit{i} = [x,y] \wedge \\& x \geq min(d_\mathit{i},H) \wedge \\&y \leq max(d_\mathit{i},H)
\end{split}
\end{equation}

, where $min(d_\mathit{i},H)$ is the lowest value of the dimension $d_\mathit{i}$ within the set $H$, and $max(d_\mathit{i},H)$ is the greatest value of the dimension $d_\mathit{i}$ within the set $H$.
\end{definition}

\begin{figure} [ht] 
	\centerline{\includegraphics[width=0.8\columnwidth]{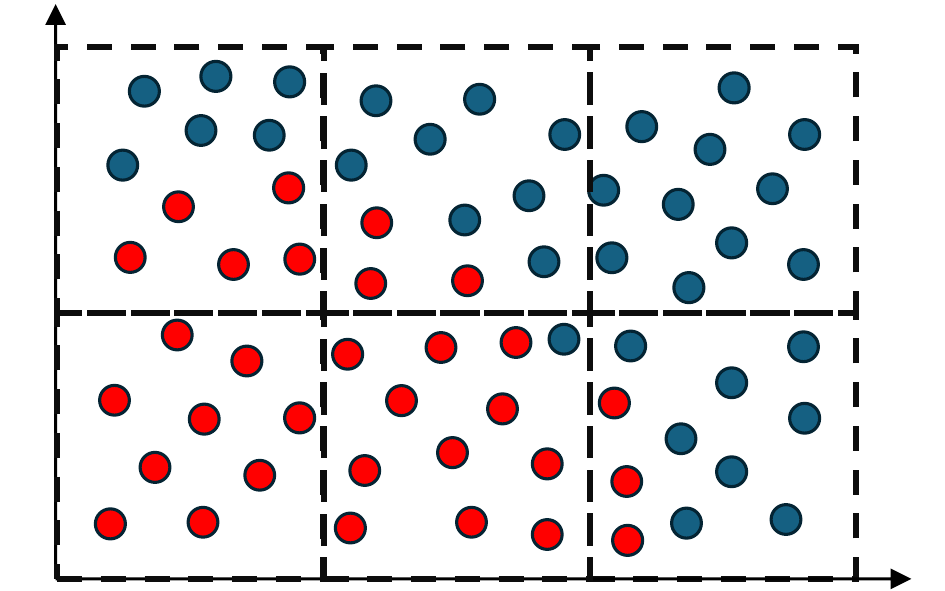}}
	\caption{Representation of a 2D space, with points from two classes (blue and red dots), split in 6 \emph{spatial partitions}.}
	\label{fig:SpatialPartitionExample}
\end{figure}

Thus, a \emph{spatial partition} of the spatial region that contains a given set of objects $H$ is a \emph{hyperrectangle} in the spatial region containing the objects in $H$. It is defined by a set of intervals, one for each dimension $d_\mathit{i} \in D$. Figure \ref{fig:SpatialPartitionExample} presents an example of a dataset with 100 data objects in a 2D space with 12 spatial partitions. 

The PSASA algorithm is formalized in Algorithm \ref{alg:PSASA}. Firstly, it splits the whole dataset into a set of spatial partitions. In the second step, for each spatial partition identified in the previous step, it identifies the set of instances of each class that belongs to that partition and produces a prototype for each of these classes. Thus, each prototype produced in this step can be viewed as the \emph{centroid} of each set of instances with the same class that is contained in each spatial partition. It is important to notice that the set of spatial partitions defined in the first step can be viewed as a \emph{spatial abstraction} of the whole dataset and that such abstraction supports the efficient creation of the set of prototypes produced by the approach. Besides that, each prototype can be viewed as an abstraction of a given set of instances in each spatial partition. Once this set of prototypes is created, it can be used as input for some conventional prototype selection approach, which can produce a smaller set of prototypes in a subsequent step.     


The PSASA algorithm takes as input a set of data objects $T$ and a value\footnote{Notice that we are assuming in this paper that the set of \emph{natural numbers} $(\mathbb{N}$) is the set of non-negative integers and, due to this, it includes zero. When we are referring to the set of natural numbers excluding zero, we use $\mathbb{N}^{*}$.} $n \in \mathbb{N}^{*}$, which determines the number of intervals in which each dimension of the dataset will be divided. In a more detailed way, firstly, the PSASA algorithm initializes $P$ as an empty set. After, it determines the set $R$ of sets of objects within $T$, such that each set $r_\mathit{i} \in R$ is a set of objects contained within a specific \emph{spatial partition} of the spatial region that contains the objects in $T$. The set $R$ is generated by the function $partitioning$, represented in Algorithm \ref{alg:partitioning}. In the next step, for each region $r_\mathit{i} \in R$, the algorithm: (I) identifies the set $I_\mathit{l}$ of instances (within $r_\mathit{i}$) of each class label $l \in L$, (II) creates a prototype $p_\mathit{l}$ (using the function $extractsPrototype$) that abstracts the information of the instances in $I_\mathit{l}$, and includes the prototype $p_\mathit{l}$ in the resulting set $P$, which is returned by the algorithm.
	
\begin{algorithm}
    \caption{PSASA algorithm}\label{alg:PSASA}
    \begin{algorithmic}[1]
        \Procedure{PSASA}{$T,n$}
            \State $P\gets \emptyset$
            \State $R \gets partitioning(T,n)$
            \ForAll{ $r_\mathit{i} \in R$}
                \ForAll{ $l \in L$}
                    \State $I_\mathit{l} \gets$ the set of instances in $r_\mathit{i}$ whose class label is $l$
		    \State $p_\mathit{l} \gets$ extractsPrototype($I_\mathit{l}$)
		    \State $P \gets P \cup \{p_\mathit{i}\}$
                \EndFor
            \EndFor
            \State \textbf{return} $P$
            \EndProcedure
    \end{algorithmic}
\end{algorithm}

The function \emph{partitioning}, proposed in our previous work \cite{carbonera2018efficient}, on the other hand, is formalized by Algorithm \ref{alg:partitioning}. This algorithm takes as input a set of instances $H$ and a number $n \in \mathbb{N}^{*}$ of intervals in which each dimension will be split. The algorithm considers the same number of intervals for every dimension. The algorithm identifies a set $R$, where each $r_\mathit{i} \in R$ is the set of instances contained in some \emph{spatial partition} of the spatial region that contains $H$. Firstly, the algorithm defines $R$ as an empty set, and then, for each dimension $d_\mathit{j} \in D$, it defines $DRange$, which is the range of the dimension $d_\mathit{j}$.  $DRange$ is defined as the absolute difference between the highest and the lowest value of $d_\mathit{j}$ in $H$. Besides that, the algorithm also defines $range_\mathit{j}$, which is the range of an interval of $d_\mathit{j}$, as $\frac{DRange}{n}$. The algorithm considers $region$ as a hash table whose keys are $m$-tuples in the form of $(x_\mathit{1},x_\mathit{2},...,x_\mathit{m})$, where each $x_\mathit{j} \leq n$ identify one of the intervals of the $j-th$ dimension in $D$. For each key $x$, $region$ stores a set of objects, in a way that $region[x] \subseteq H$. Thus, each key of $region$ can be viewed as the identification of some \emph{spatial partition}. Thus, $region[x]$ represents the set of objects located within the spatial partition identified by $x$. After, for each object $o_\mathit{i} \in H$, the algorithm:

\begin{enumerate}
    \item Considers $x$ as an empty $|D|$-tuple.
    \item For each dimension $d_\mathit{j} \in D$, it calculates the value of $x_\mathit{j}$ that identifies the interval that contains the value $o_\mathit{ij}$, such that:
    \begin{equation}
    x_\mathit{j} \gets  \lfloor \frac{o_\mathit{ij}-min(d_\mathit{i},H)}{range_\mathit{i}} \rfloor.
    \end{equation}
    Thus, the m-tuple $x$ identifies the spatial partition that contains $o_\mathit{i}$.
    \item Includes $o$ as an element of $region[x]$.
\end{enumerate}

Finally, for each key $x$ of $region$, the algorithm includes the set $region[x]$ in $R$ as an element. Thus, each element of $R$ is a set of objects (that can belong to different classes) located in some spatial partition identified by the algorithm. Finally, the algorithm returns $R$. Notice that since the algorithm is focused on objects, it computes only the spatial partitions that contain objects and returns only \emph{non-empty} regions. 

\begin{algorithm}
    \caption{partitioning function}\label{alg:partitioning}
    \begin{algorithmic}[1]
        \Procedure{partitioning}{$H,n$}
            \State $R\gets \emptyset$
            \State Let $range_\mathit{j}$ be the range of an interval of the dimension $d_\mathit{j} \in D$.
           \ForAll{ $d_\mathit{j} \in D$}
                \State $DRange \gets abs( max(d_\mathit{j},H) - min(d_\mathit{j},H) )$
		      \State $range_\mathit{j} \gets $ \large{$\frac{DRange}{n}$}
            \EndFor
            \State Let $region$ be a hash table whose keys are $m$-tuples in the form of $(x_\mathit{1},x_\mathit{2},...,x_\mathit{m})$, where each $x_\mathit{j}$ identifies one of the intervals of the $j- th$ dimension within $D$. 
            Thus, for each key $x$, $region$ stores a set of objects, such that $region[x] \subseteq H$.
           \ForAll{ $o_\mathit{i} \in H$}
                \State Let $x$ be an empty $m$-tuple.
		      \ForAll{ $d_\mathit{j} \in D$}
                    \State $x_\mathit{j} \gets $ \large{$\lfloor \frac{o_\mathit{ij}-                           min(d_\mathit{j},H)}{range_\mathit{j}} \rfloor$}
                \EndFor
                \State $region[x] \gets region[x] \cup \{o_\mathit{i}\}$
            \EndFor
            \ForAll{ key $x$ of $region$}
                \State $R$ includes $region[x]$ as its element\;
            \EndFor
            \State \textbf{return} $R$
            \EndProcedure
    \end{algorithmic}
\end{algorithm}

The function \emph{extractsPrototype}, adopted by the Algorithm \ref{alg:PSASA}, takes as input a set of instances $W \subseteq T$ and produces a $m$-tuple that represents the centroid of the instances in $W$. This is the same strategy used by \cite{carbonera2017ICTAI} for extracting prototypes, for example.

Notice that Algorithm \ref{alg:partitioning} assumes that the spatial region that includes all the elements in $H$ is divided in a set of \emph{non-overlapping} spatial partitions, which covers all the set $H$.

Moreover, since the notion of \emph{spatial partition} defined in this work can be applied only to datasets with quantitative (numerical) dimensions, the PSASA algorithm inherits the same limitation and can be applied only to numerical datasets. 
	
Besides that, it is important to notice that each of the steps of the algorithm has, at most, a time complexity that is \emph{linear} on the number of instances. Therefore, the PSASA algorithm is able to produce a set of candidate prototypes in a very \emph{efficient way}.

\begin{figure}[!ht] 
	\centerline{\includegraphics[width=0.8\columnwidth]{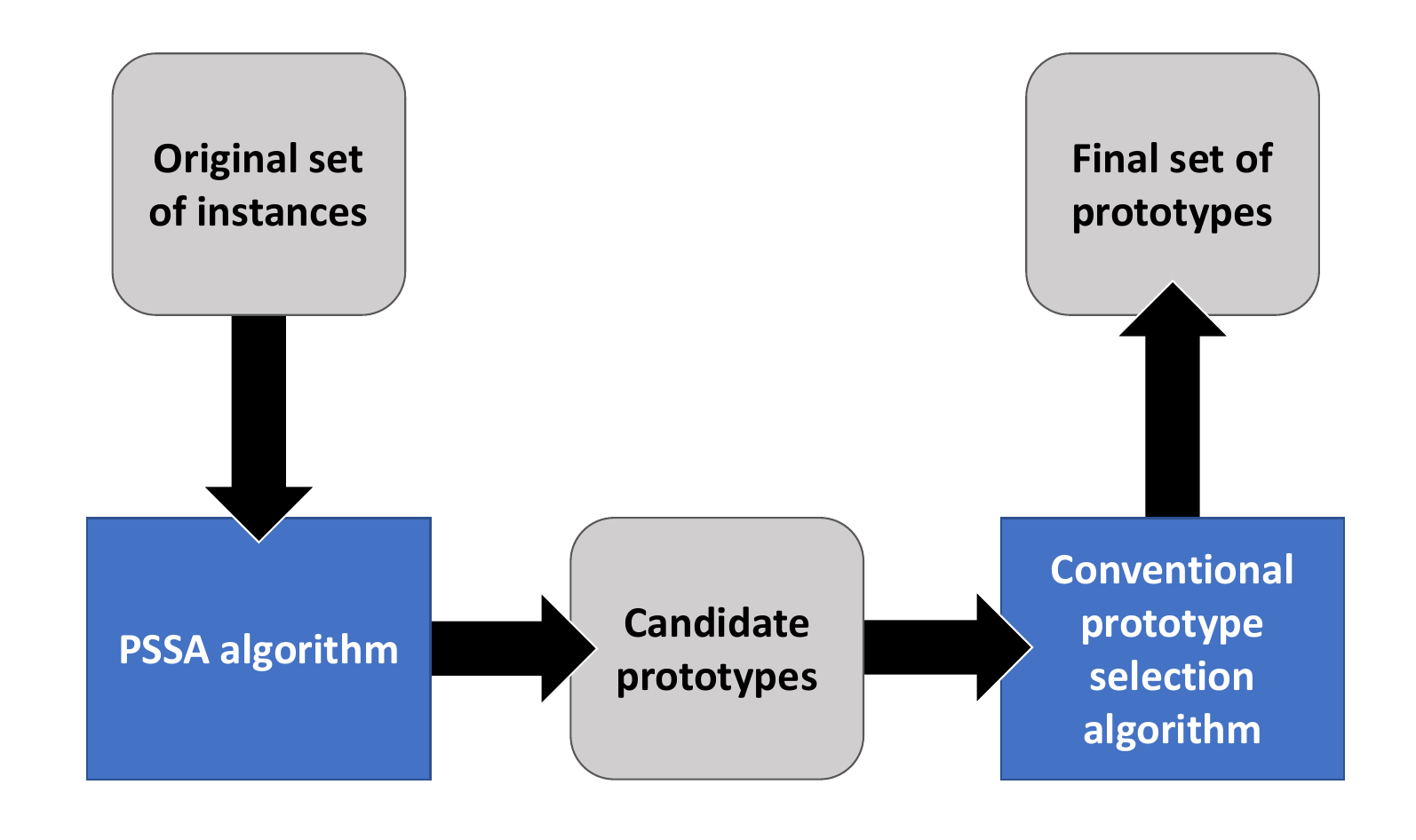}}
	\caption{General schema of a pipeline that uses the PSASA algorithm for speeding up some conventional prototype selection algorithm.}
	\label{fig:Pipeline}
\end{figure}

Finally, as previously mentioned, the PSASA algorithm can be used in \emph{pipelines} of prototype selection as a pre-stage that produces a set of candidate prototypes, which can be refined by some conventional prototype selection algorithm. Figure \ref{fig:Pipeline} represents the general schema of such pipelines. Notice that the output of the pipelines developed with the PSASA algorithm is always a set of \emph{synthetic} data points, even when we adopt, in the final stage of the pipeline, a prototype selection algorithm that was originally designed for selecting a subset of the original dataset. This happens because the PSASA algorithm creates synthetic data points that abstract the information of subsets of the original dataset. However, the algorithm could also be modified to select a subset of actual instances instead of synthetic prototypes. In order to do this, in the last step of PSASA it would be necessary to find the nearest actual instances of the prototypes that were generated by the algorithm in its current version.

\section{Experiments} \label{sec:experiments}
	
For evaluating our approach, we selected 5 important prototype selection algorithms\footnote{All algorithms were implemented by the authors.} available in the literature: DROP3, ENN, ICF, LSBo and LSSm. Notice that these algorithms were proposed for selecting \emph{actual instances} (instead of synthetic ones) within the original dataset. Then, we developed a modified version of each algorithm, adopting the PSASA algorithm as a pre-step in a pipeline that follows the general schema presented in Figure \ref{fig:Pipeline}. The resulting \emph{enhanced algorithms} were called, respectively: Fast DROP3, Fast ENN, Fast ICF, Fast LSBo and Fast LSSm. These algorithms produce as output sets of synthetic data points that are created by the PSASA algorithm. The modified versions of the algorithms were compared with their original versions in a classification task, considering 14 well-known datasets with numerical dimensions: cardiotocography, diabetes, E. Coli, glass, heart-statlog, ionosphere, iris, landsat, letter, optdigits, parkinson, segment, spambase, and wine. All datasets were obtained from the UCI Machine Learning Repository\footnote{\url{http://archive.ics.uci.edu/ml/}}. Table \ref{tab:datasets} presents the details of the datasets that were used.
	
	\begin{table}[h]\centering 
		\caption{Details of the datasets used in the evaluation process.}\label{tab:datasets}
			\begin{tabular}{l|*{4}{c}}
				\hline
				Dataset		  					  & Instances 		& Attributes	   & Classes	   \\\hline\hline
				{\bf Cardiotocography}            & 2126            & 21               & 10            \\
				{\bf Diabetes}            & 768            & 9               & 2            \\
				{\bf E. Coli}                     & 336             & 8                & 8             \\
				{\bf Glass}                       & 214             & 10               & 7             \\
				{\bf Heart-statlog}               & 270             & 14               & 2             \\
				{\bf Ionosphere}                  & 351             & 35               & 2             \\
				{\bf Iris}                        & 150             & 5                & 3             \\
				{\bf Landsat}                     & 4435            & 37               & 6             \\
				{\bf Letter}                      & 20000           & 17               & 26            \\
				{\bf Optdigits}                   & 11240             & 65                & 10             \\
				{\bf Parkinson}                   & 195             & 23                & 2             \\
				{\bf Spambase}                     & 9544            & 58               & 2             \\
				{\bf Segment}                     & 2310            & 20               & 7             \\
				{\bf Wine}                        & 178             & 14               & 3             \\ \hline
			\end{tabular}
	\end{table}

We adopted two measures to evaluate the performance of the algorithms: \emph{accuracy} and \emph{reduction}. Following \cite{LEYVA:2015}, we assume $accuracy = |Sucess(Test)|/|Test|$ and $reduction = (|T|-|S|)/|T|$, where $Test$ is a set of instances that were selected for being tested in a classification task, and $|Success(Test)|$ is the number of instances in $Test$ correctly classified in the classification task.
	
The classification \emph{accuracy} of new instances in each respective dataset was evaluated with two different classifiers: KNN and SVM. For the KNN classifier, we considered $k=3$, as assumed in \cite{LEYVA:2015}. For the SVM, following \cite{anwar2015instance}, we considered the implementation provided by Weka $3.8$, with the standard parametrization ( $c=1.0$, $toleranceParameter= 0.001$, $epsilon=1.0E-12$, using a polynomial kernel and a multinomial logistic regression model with a ridge estimator as calibrator).

Following the procedure adopted in \cite{CARBONERA2015ICTAI}, the accuracy and reduction were evaluated in an \emph{n-fold cross-validation} scheme, where $n=10$. 
	
In all experiments, following the procedure adopted by \cite{CARBONERA2015ICTAI}, we adopted $k = 3$ for DROP3, ENN, ICF, and for their modified versions. For the algorithms enhanced by PSASA, we adopted $n=5$ since, according to our experiments, this value of $n$ provides a good balance between accuracy, reduction, and running time. Besides that, for the algorithms that adopt some distance (dissimilarity) function, we considered the standard \emph{Euclidean} distance.

Tables \ref{table:accuracyKNN} and \ref{table:accuracySVM} show that the accuracy achieved by each algorithm adopting the PSASA algorithm as a pre-step is similar to the accuracy achieved by the original algorithm in most of the datasets. This similarity can also be observed in the average of the accuracy of each algorithm, considering all the datasets. This behavior can be observed for both the considered classifiers, KNN and SVM. Notice also that in some cases, the performance of the enhanced algorithm is greater than the performance achieved by its original version. Thus, the results suggest that, with the considered parameters, in most of the cases, the set of prototypes created by the PSASA algorithm is able to preserve the information that is contained in the original set of instances to a degree that is comparable to the information preserved by the original versions of the algorithms.

\begin{table*}[!ht]
\centering
\caption{\emph{Accuracy} achieved by the training set produced by each algorithm, for each dataset, adopting a KNN classifier.}
\label{table:accuracyKNN}
\resizebox{\textwidth}{!}{%
\begin{tabular}{|l|c|c|c|c|c|c|c|c|c|c|c|}
\hline
\multirow{2}{*}{\textbf{Dataset}} & \multicolumn{10}{c|}{\textbf{Algorithm}}                                                                                                                                                                                                                                                                                                                                                                 & \multicolumn{1}{l|}{\multirow{2}{*}{\textbf{Average}}} \\ 
                                  & \textbf{DROP3} & \textbf{\begin{tabular}[c]{@{}c@{}}Fast\\ DROP3\end{tabular}} & \textbf{ENN}  & \textbf{\begin{tabular}[c]{@{}c@{}}Fast\\ ENN\end{tabular}} & \textbf{ICF}  & \textbf{\begin{tabular}[c]{@{}c@{}}Fast\\ ICF\end{tabular}} & \textbf{LSBo} & \textbf{\begin{tabular}[c]{@{}c@{}}Fast\\ LSBo\end{tabular}} & \textbf{LSSm} & \textbf{\begin{tabular}[c]{@{}c@{}}Fast\\ LSSm\end{tabular}} & \multicolumn{1}{l|}{}                                  \\ \hline\hline
\textbf{Cardiotocography}         & 0.63           & 0.59                                                          & 0.64          & 0.64                                                        & 0.57          & 0.58                                                        & 0.55          & 0.54                                                         & \textbf{0.67} & 0.66                                                         & 0.61                                                   \\ \hline
\textbf{Diabetes}                 & 0.72           & 0.67                                                          & \textbf{0.72} & 0.70                                                        & \textbf{0.72} & 0.64                                                        & 0.73          & 0.66                                                         & \textbf{0.72} & 0.69                                                         & 0.70                                                   \\ \hline
\textbf{E. Coli}                  & 0.84           & 0.83                                                          & 0.84          & 0.85                                                        & 0.79          & 0.83                                                        & 0.79          & 0.79                                                         & \textbf{0.86} & 0.85                                                         & 0.83                                                   \\ \hline
\textbf{Glass}                    & 0.63           & 0.61                                                          & 0.63          & 0.65                                                        & 0.64          & 0.66                                                        & 0.54          & 0.59                                                         & \textbf{0.71} & 0.69                                                         & 0.64                                                   \\ \hline
\textbf{Heart-statlog}            & 0.67           & 0.67                                                          & 0.64          & \textbf{0.68}                                               & 0.63          & 0.66                                                        & 0.66          & \textbf{0.68}                                                & 0.66          & 0.67                                                         & 0.66                                                   \\ \hline
\textbf{Ionosphere}               & 0.82           & 0.87                                                          & 0.83          & 0.89                                                        & 0.82          & 0.87                                                        & 0.88          & 0.90                                                         & 0.86          & \textbf{0.91}                                                & 0.87                                                   \\ \hline
\textbf{Iris}                     & 0.97           & 0.96                                                          & \textbf{0.97} & 0.95                                                        & 0.95          & 0.91                                                        & 0.95          & 0.93                                                         & 0.96          & \textbf{0.97}                                                & 0.95                                                   \\ \hline
\textbf{Landsat}                  & 0.88           & 0.89                                                          & \textbf{0.90} & 0.89                                                        & 0.83          & 0.86                                                        & 0.86          & 0.86                                                         & \textbf{0.90} & \textbf{0.90}                                                & 0.88                                                   \\ \hline
\textbf{Letter}                   & 0.88           & 0.84                                                          & 0.92          & 0.90                                                        & 0.80          & 0.77                                                        & 0.73          & 0.72                                                         & \textbf{0.93} & 0.90                                                         & 0.84                                                   \\ \hline
\textbf{Optdigits}                & 0.97           & 0.97                                                          & \textbf{0.98} & \textbf{0.98}                                               & 0.91          & 0.91                                                        & 0.91          & 0.91                                                         & \textbf{0.98} & \textbf{0.98}                                                & 0.95                                                   \\ \hline
\textbf{Parkinsons}               & 0.83           & 0.84                                                          & 0.85          & 0.85                                                        & 0.81          & 0.83                                                        & 0.85          & \textbf{0.87}                                                & 0.85          & 0.86                                                         & 0.85                                                   \\ \hline
\textbf{Segment}                  & 0.92           & 0.90                                                          & \textbf{0.94} & 0.92                                                        & 0.87          & 0.87                                                        & 0.83          & 0.84                                                         & \textbf{0.94} & \textbf{0.94}                                                & 0.90                                                   \\ \hline
\textbf{Spambase}                 & 0.79           & 0.75                                                          & 0.81          & 0.77                                                        & 0.79          & 0.76                                                        & 0.81          & 0.77                                                         & \textbf{0.82} & 0.77                                                         & 0.78                                                   \\ \hline
\textbf{Wine}                     & 0.69           & 0.68                                                          & 0.66          & 0.74                                                        & 0.66          & 0.78                                                        & 0.74          & 0.75                                                         & 0.71          & \textbf{0.79}                                                & 0.72                                                   \\ \hline\hline
\textbf{Average}                  & 0.80           & 0.79                                                          & 0.81          & 0.82                                                        & 0.77          & 0.78                                                        & 0.77          & 0.77                                                         & \textbf{0.83} & \textbf{0.83}                                                & 0.80                                                   \\ \hline
\end{tabular}%
}
\end{table*}

\begin{table*}[!ht]
\centering
\caption{\emph{Accuracy} achieved by the training set produced by each algorithm, for each dataset, adopting a SVM classifier.}
\label{table:accuracySVM}
\resizebox{\textwidth}{!}{
\begin{tabular}{|l|c|c|c|c|c|c|c|c|c|c|c|}
\hline
\multirow{2}{*}{\textbf{Dataset}}& \multicolumn{10}{c|}{\textbf{Algorithm}}                                                                                                                                                                                                                                                                                                                                                                 & \multicolumn{1}{l|}{\multirow{2}{*}{\textbf{Average}}} \\ 
                                  & \textbf{DROP3} & \textbf{\begin{tabular}[c]{@{}c@{}}Fast\\ DROP3\end{tabular}} & \textbf{ENN}  & \textbf{\begin{tabular}[c]{@{}c@{}}Fast\\ ENN\end{tabular}} & \textbf{ICF}  & \textbf{\begin{tabular}[c]{@{}c@{}}Fast\\ ICF\end{tabular}} & \textbf{LSBo} & \textbf{\begin{tabular}[c]{@{}c@{}}Fast\\ LSBo\end{tabular}} & \textbf{LSSm} & \textbf{\begin{tabular}[c]{@{}c@{}}Fast\\ LSSm\end{tabular}} &                                   \\ \hline\hline
\textbf{Cardiotocography}         & 0.64           & 0.65                                                          & \textbf{0.67} & \textbf{0.67}                                                        & 0.64          & 0.65                                                        & 0.62          & 0.63                                                         & \textbf{0.67} & \textbf{0.67}                                                & 0.65                              \\ \hline
\textbf{Diabetes}                 & 0.75           & 0.76                                                          & \textbf{0.77} & \textbf{0.77}                                               & 0.76          & 0.75                                                        & 0.75          & 0.76                                                         & \textbf{0.77} & 0.76                                                         & 0.76                              \\ \hline
\textbf{E. Coli}                  & 0.81           & 0.81                                                          & 0.82          & 0.82                                                        & 0.78          & 0.79                                                        & 0.74          & 0.73                                                         & \textbf{0.83} & \textbf{0.83}                                                & 0.80                              \\ \hline
\textbf{Glass}                    & 0.47           & 0.50                                                          & 0.49          & 0.53                                                        & 0.49          & 0.54                                                        & 0.42          & 0.46                                                         & \textbf{0.55} & 0.48                                                         & 0.49                              \\ \hline
\textbf{Heart-statlog}            & 0.81           & \textbf{0.86}                                                 & 0.83          & 0.85                                                        & 0.79          & 0.81                                                        & 0.81          & 0.82                                                         & 0.84          & 0.84                                                         & 0.83                              \\ \hline
\textbf{Ionosphere}               & 0.81           & 0.81                                                          & 0.87          & 0.86                                                        & 0.58          & 0.76                                                        & 0.45          & 0.56                                                         & \textbf{0.88} & \textbf{0.88}                                                & 0.75                              \\ \hline
\textbf{Iris}                     & 0.94           & 0.90                                                          & \textbf{0.96} & 0.93                                                        & 0.73          & 0.72                                                        & 0.47          & 0.65                                                         & \textbf{0.96} & 0.94                                                         & 0.82                              \\ \hline
\textbf{Landsat}                  & 0.86           & 0.86                                                          & \textbf{0.87} & \textbf{0.87}                                               & 0.85          & 0.85                                                        & 0.85          & 0.85                                                         & \textbf{0.87} & \textbf{0.87}                                                & 0.86                              \\ \hline
\textbf{Letter}                   & 0.80           & 0.78                                                          & \textbf{0.84} & 0.81                                                        & 0.75          & 0.74                                                        & 0.73          & 0.73                                                         & \textbf{0.84} & 0.81                                                         & 0.78                              \\ \hline
\textbf{Optdigits}                & 0.98           & 0.98                                                          & 0.98          & 0.98                                                        & 0.97          & 0.97                                                        & 0.98          & 0.98                                                         & \textbf{0.99} & 0.98                                                         & 0.98                              \\ \hline
\textbf{Parkinsons}               & 0.83           & 0.83                                                          & 0.86          & \textbf{0.87}                                               & 0.82          & 0.81                                                        & 0.81          & 0.81                                                         & 0.86          & \textbf{0.87}                                                & 0.84                              \\ \hline
\textbf{Segment}                  & 0.91           & 0.89                                                          & \textbf{0.92} & 0.90                                                        & 0.91          & 0.88                                                        & 0.80          & 0.77                                                         & 0.91          & 0.91                                                         & 0.88                              \\ \hline
\textbf{Spambase}                 & 0.90           & 0.87                                                          & \textbf{0.90} & 0.85                                                        & \textbf{0.90} & 0.86                                                        & \textbf{0.90} & 0.85                                                         & \textbf{0.90} & 0.84                                                         & 0.88                              \\ \hline
\textbf{Wine}                     & 0.93           & 0.95                                                          & 0.95          & 0.96                                                        & 0.94          & 0.94                                                        & 0.96          & 0.94                                                         & \textbf{0.97} & \textbf{0.97}                                                & 0.95                              \\ \hline\hline
\textbf{Average}                  & 0.82           & 0.82                                                          & 0.84          & 0.83                                                        & 0.78          & 0.79                                                        & 0.74          & 0.75                                                         & \textbf{0.85} & 0.83                                                         & 0.80                              \\ \hline
\end{tabular}%
}
\end{table*}

\begin{figure*} [!ht]
	\centerline{\includegraphics[width=0.9\textwidth]{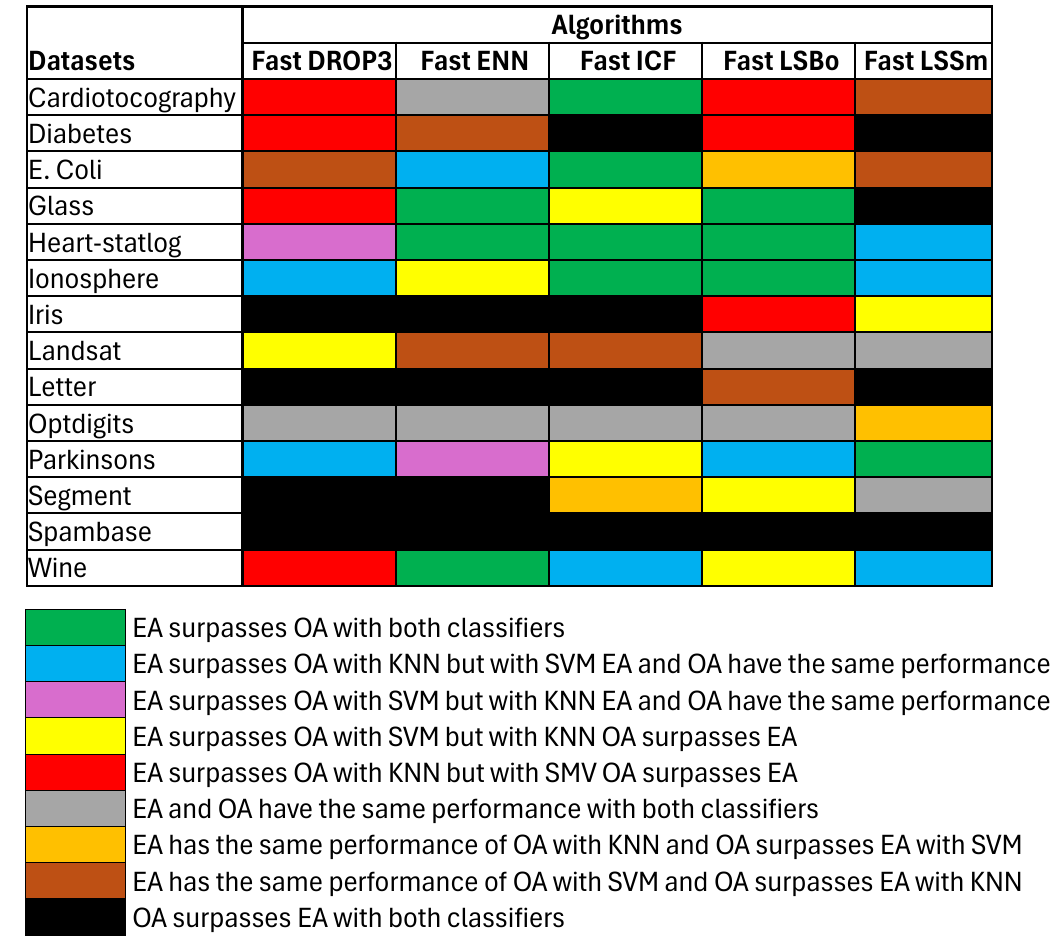}}
	\caption{Detailed qualitative comparison among the conventional prototype selection algorithms and their enhanced versions. OA means \emph{Original algorithm} and EA means \emph{Enhanced Algorithm}}
	\label{fig:qualitativeDetailed}
\end{figure*}

\begin{figure*} [!ht]
	\centerline{\includegraphics[width=0.9\textwidth]{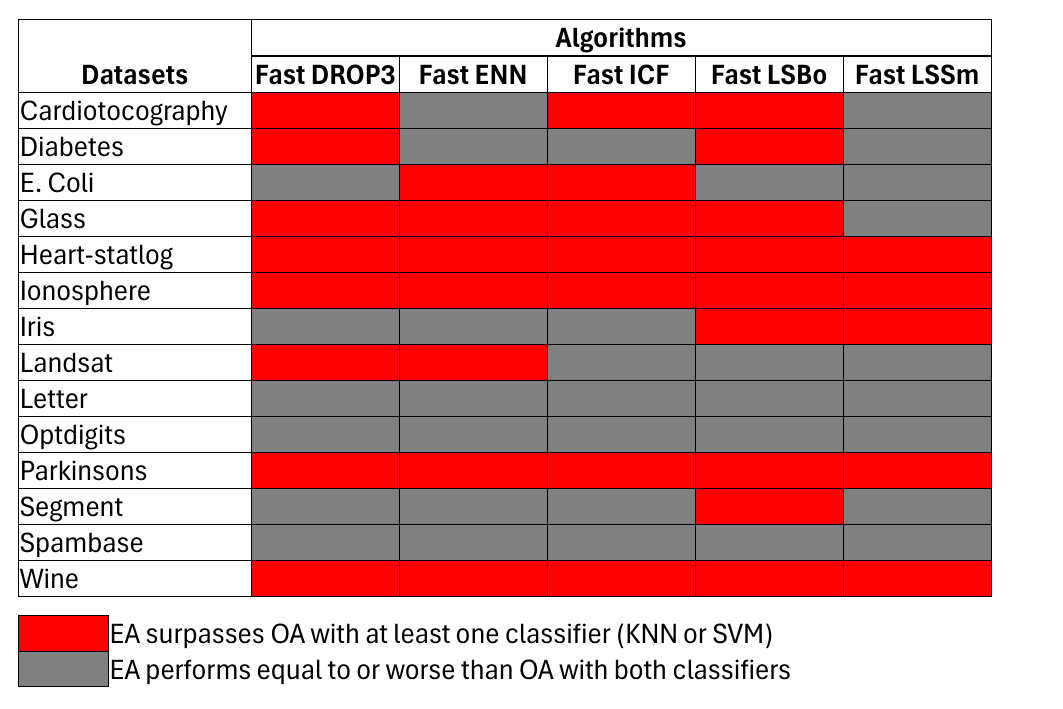}}
	\caption{Abstract qualitative comparison among the conventional prototype selection algorithms and their enhanced versions. OA means \emph{Original algorithm} and EA means \emph{Enhanced Algorithm}}
	\label{fig:qualitativeAbstract}
\end{figure*}

In order to facilitate the qualitative comparison of the performance of the conventional prototype selection algorithms and their enhanced versions, we propose the visualization presented in Figure \ref{fig:qualitativeDetailed}, that summarizes the results of Tables \ref{table:accuracyKNN} and \ref{table:accuracySVM}. This visualization allows us to identify cases where the performance of the enhanced algorithms surpasses or is surpassed by or has the same performance as the original algorithms with each classifier. On the other hand, Figure \ref{fig:qualitativeAbstract} provides a more abstract view of the same information. This visualization emphasizes the cases in which the enhanced version of the algorithms surpasses the original algorithm with at least one classifier (KNN or SVM or both), and the cases in which the enhanced algorithm performs equal to or worse than the original algorithm considering both classifiers. This analysis shows that the enhanced algorithms surpass the original ones in 36 cases, corresponding to $51\%$ of the 70 considered cases. This is an interesting result, since the PSASA was not designed to improve classification accuracy, but to speed up the conventional prototype selection algorithms. Figure \ref{fig:qualitativeAbstract} also allows us to identify datasets in which the enhanced version always surpasses the corresponding original algorithm (with at least one classifier), such as: Heart-statlog, Ionosphere, Parkinsons, and Wine. On the other hand, Figure \ref{fig:qualitativeAbstract} also highlights datasets in which the enhanced version was not able to surpass the original algorithm in any case, such as: Letter, Optdigits, and Spambase. Future works can investigate which properties of these datasets cause these observed patterns.

It is important to notice that the absence of improvement in accuracy cannot be considered as a negative result in this context if the approach provides gains in reduction rate and running time. The next results show that this is exactly the case.

Regarding the reduction rate, Table \ref{table:reduction} shows that the modified versions of the considered algorithms achieved the best results in most of the datasets when compared with the respective original versions of the algorithm. This behavior can be observed also when we analyze the average reduction rate of each algorithm, considering all the datasets. In most of cases, there are big differences between the average reduction achieved by an algorithm and its faster version. This difference in performance is particularly significant in the case of \emph{Fast LSSm} and \emph{Fast ENN}, when compared with the reduction achieved by the original versions of these algorithms.

\begin{table*}[!ht]
\centering
\caption{\emph{Reduction rate} achieved by each algorithm, for each dataset.}
		\label{table:reduction}
		\resizebox{\textwidth}{!}{%
\begin{tabular}{|l|c|c|c|c|c|c|c|c|c|c|c|}
\hline
\multirow{2}{*}{\textbf{Dataset}} & \multicolumn{10}{c|}{\textbf{Algorithm}}                                                                                                                                                                                                                                                                                                                                                                & \multirow{2}{*}{\textbf{Average}} \\ 
                                  & \textbf{DROP3} & \textbf{\begin{tabular}[c]{@{}c@{}}Fast\\ DROP3\end{tabular}} & \textbf{ENN} & \textbf{\begin{tabular}[c]{@{}c@{}}Fast\\ ENN\end{tabular}} & \textbf{ICF}  & \textbf{\begin{tabular}[c]{@{}c@{}}Fast\\ ICF\end{tabular}} & \textbf{LSBo} & \textbf{\begin{tabular}[c]{@{}c@{}}Fast\\ LSBo\end{tabular}} & \textbf{LSSm} & \textbf{\begin{tabular}[c]{@{}c@{}}Fast\\ LSSm\end{tabular}} &                                   \\ \hline\hline
\textbf{Cardiotocography}         & 0.70           & 0.77                                                          & 0.32         & 0.50                                                        & 0.71          & \textbf{0.78}                                               & 0.69          & 0.72                                                         & 0.14          & 0.31                                                         & 0.56                              \\ \hline
\textbf{Diabetes}                 & 0.77           & 0.84                                                          & 0.31         & 0.57                                                        & 0.85          & \textbf{0.86}                                               & 0.76          & 0.78                                                         & 0.13          & 0.42                                                         & 0.63                              \\ \hline
\textbf{E. Coli}                  & 0.72           & 0.82                                                          & 0.17         & 0.54                                                        & 0.87          & \textbf{0.88}                                               & 0.83          & 0.85                                                         & 0.09          & 0.46                                                         & 0.62                              \\ \hline
\textbf{Glass}                    & 0.75           & \textbf{0.86}                                                 & 0.35         & 0.65                                                        & 0.69          & 0.78                                                        & 0.70          & 0.81                                                         & 0.13          & 0.50                                                         & 0.62                              \\ \hline
\textbf{Heart-statlog}            & 0.74           & 0.73                                                          & 0.35         & 0.31                                                        & 0.78          & \textbf{0.79}                                               & 0.67          & 0.69                                                         & 0.15          & 0.16                                                         & 0.54                              \\ \hline
\textbf{Ionosphere}               & 0.86           & 0.82                                                          & 0.15         & 0.24                                                        & \textbf{0.96} & 0.91                                                        & 0.81          & 0.86                                                         & 0.04          & 0.18                                                         & 0.58                              \\ \hline
\textbf{Iris}                     & 0.70           & 0.88                                                          & 0.04         & 0.67                                                        & 0.61          & 0.89                                                        & 0.92          & \textbf{0.95}                                                & 0.05          & 0.68                                                         & 0.64                              \\ \hline
\textbf{Landsat}                  & 0.72           & 0.74                                                          & 0.10         & 0.15                                                        & \textbf{0.91} & \textbf{0.91}                                               & 0.88          & 0.88                                                         & 0.05          & 0.10                                                         & 0.54                              \\ \hline
\textbf{Letter}                   & 0.68           & 0.78                                                          & 0.05         & 0.42                                                        & 0.80          & 0.85                                                        & 0.84          & \textbf{0.86}                                                & 0.04          & 0.40                                                         & 0.57                              \\ \hline
\textbf{Optdigits}                & 0.72           & 0.71                                                          & 0.01         & 0.02                                                        & \textbf{0.93} & 0.92                                                        & 0.92          & 0.90                                                         & 0.02          & 0.02                                                         & 0.52                              \\ \hline
\textbf{Parkinsons}               & 0.71           & 0.70                                                          & 0.15         & 0.14                                                        & 0.75          & 0.75                                                        & 0.87          & \textbf{0.88}                                                & 0.12          & 0.13                                                         & 0.52                              \\ \hline
\textbf{Segment}                  & 0.68           & 0.87                                                          & 0.05         & 0.64                                                        & 0.79          & 0.91                                                        & 0.90          & \textbf{0.94}                                                & 0.05          & 0.63                                                         & 0.65                              \\ \hline
\textbf{Spambase}                 & 0.74           & 0.94                                                          & 0.19         & 0.83                                                        & 0.79          & \textbf{0.97}                                               & 0.82          & 0.96                                                         & 0.10          & 0.82                                                         & 0.72                              \\ \hline
\textbf{Wine}                     & 0.80           & 0.70                                                          & 0.30         & 0.23                                                        & \textbf{0.82} & 0.80                                                        & 0.75          & 0.78                                                         & 0.11          & 0.10                                                         & 0.54                              \\ \hline\hline
\textbf{Average}                  & 0.73           & 0.80                                                          & 0.18         & 0.42                                                        & 0.80          & \textbf{0.86}                                               & 0.81          & 0.85                                                         & 0.09          & 0.35                                                         & 0.59                              \\ \hline
\end{tabular}%
}
\end{table*}
	
\begin{figure*} [!ht]
	\centerline{\includegraphics[width=0.9\textwidth]{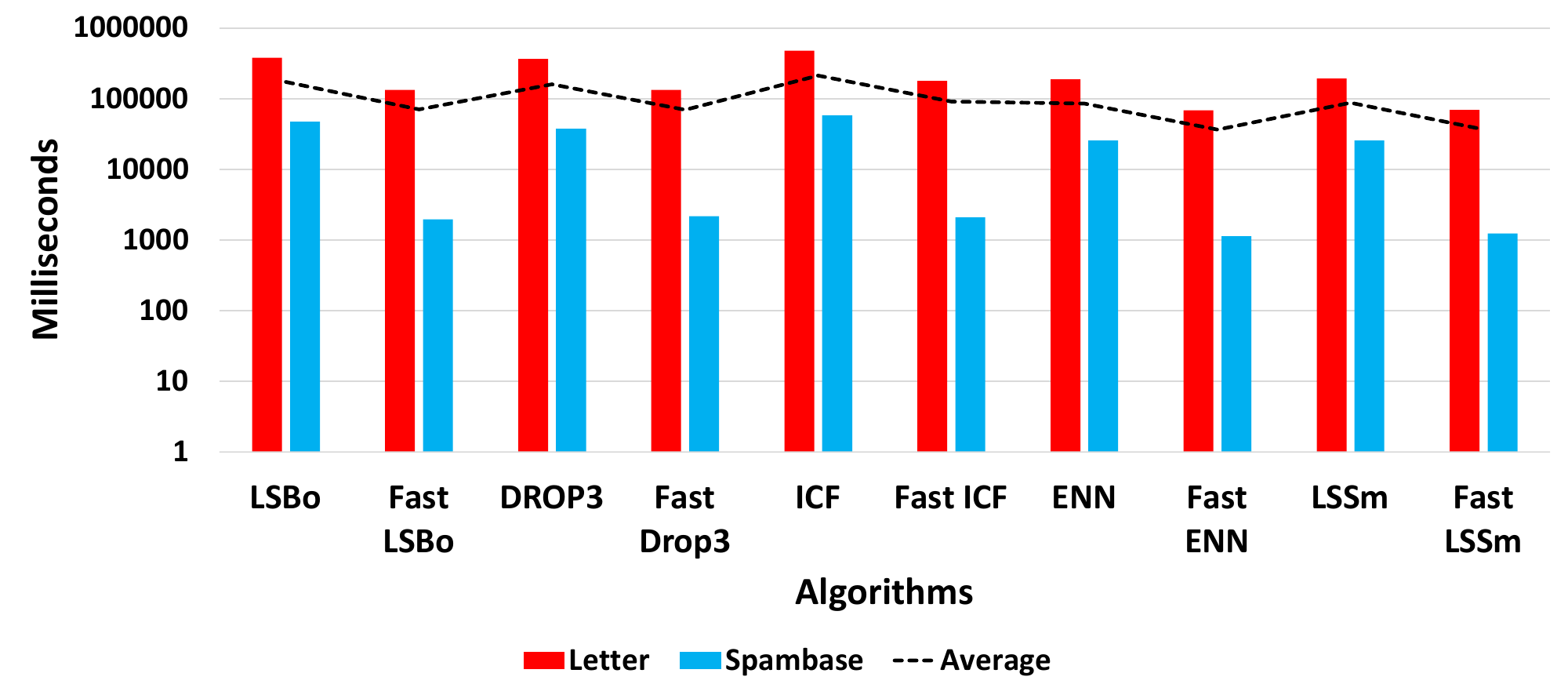}}
	\caption{Comparison of the running times of the 10 prototype selection algorithms, considering two datasets. Notice that the time axis uses a logarithmic scale.}
	\label{fig:times}
\end{figure*}

We also compared of the running times of the prototype selection algorithms considered in our experiments. In this comparison, we applied the 10 prototype selection algorithms (the 5 original algorithms and their versions with the PSASA algorithm) to reduce 2 of the biggest datasets considered in our tests: \emph{letter} and \emph{spambase}. These datasets were selected due to their sizes and also because they have different characteristics. The \emph{letter} dataset has a big amount of instances, which are described with a relatively small number of features and classified according to a big set of classes. On the other hand, the \emph{spambase} dataset has half the number of instances of the \emph{letter} dataset, but has a larger set of features and a smaller set of classes. In this experiment, we adopted the same parametrizations that were adopted in the first experiment. We performed the experiments in an Intel\textsuperscript{\textregistered} Core\textsuperscript{TM} i5-5200U laptop with a 2.2 GHz CPU and 8 GB of RAM. Figure \ref{fig:times} shows that considering these datasets, the running times achieved by the algorithms that adopt the PSASA algorithm as a pre-step are lower than the running times of their respective original versions. This suggests that the PSASA algorithm provides an effective speed-up to conventional prototype selection algorithms. The results also show that the approach provided a higher speed-up in the \emph{spambase} dataset. This is expected because, since is has a lower number of classes in comparison with \emph{letter} dataset, it produces a lower number of prototypes in each spatial partition, saving processing time. This difference agrees with the difference in reduction rates (Table \ref{table:reduction}) between each algorithm and its enhanced version, regarding the \emph{letter} and \emph{spambase} datasets.

\begin{figure*}[!ht] 
	\centerline{\includegraphics[width=0.9\textwidth]{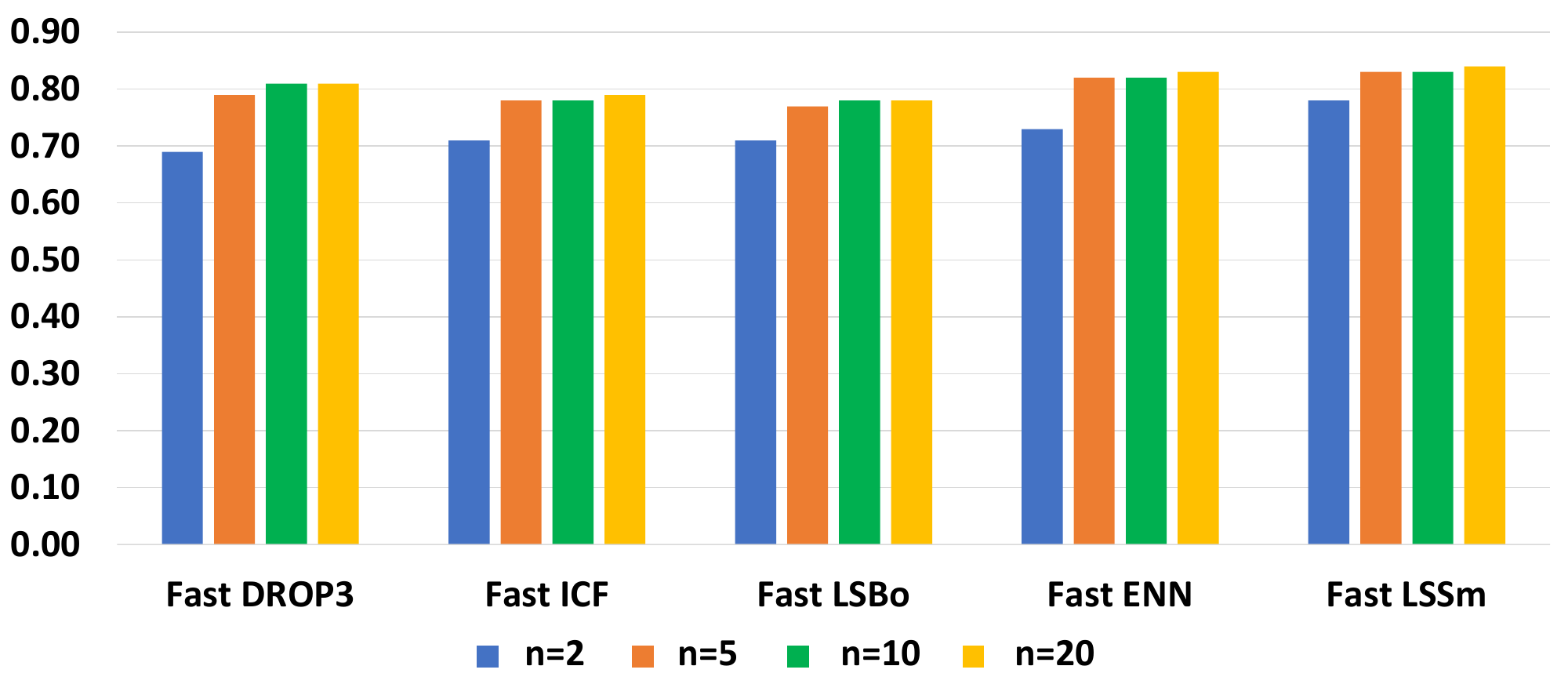}}
	\caption{Average accuracy (considering the accuracy in all datasets), for each value of $n$, of the 5 algorithms that use the PSASA algorithm as a pre-step.}
	\label{fig:accuracyVaryingN}
\end{figure*}

\begin{figure*}[!ht] 
	\centerline{\includegraphics[width=0.9\textwidth]{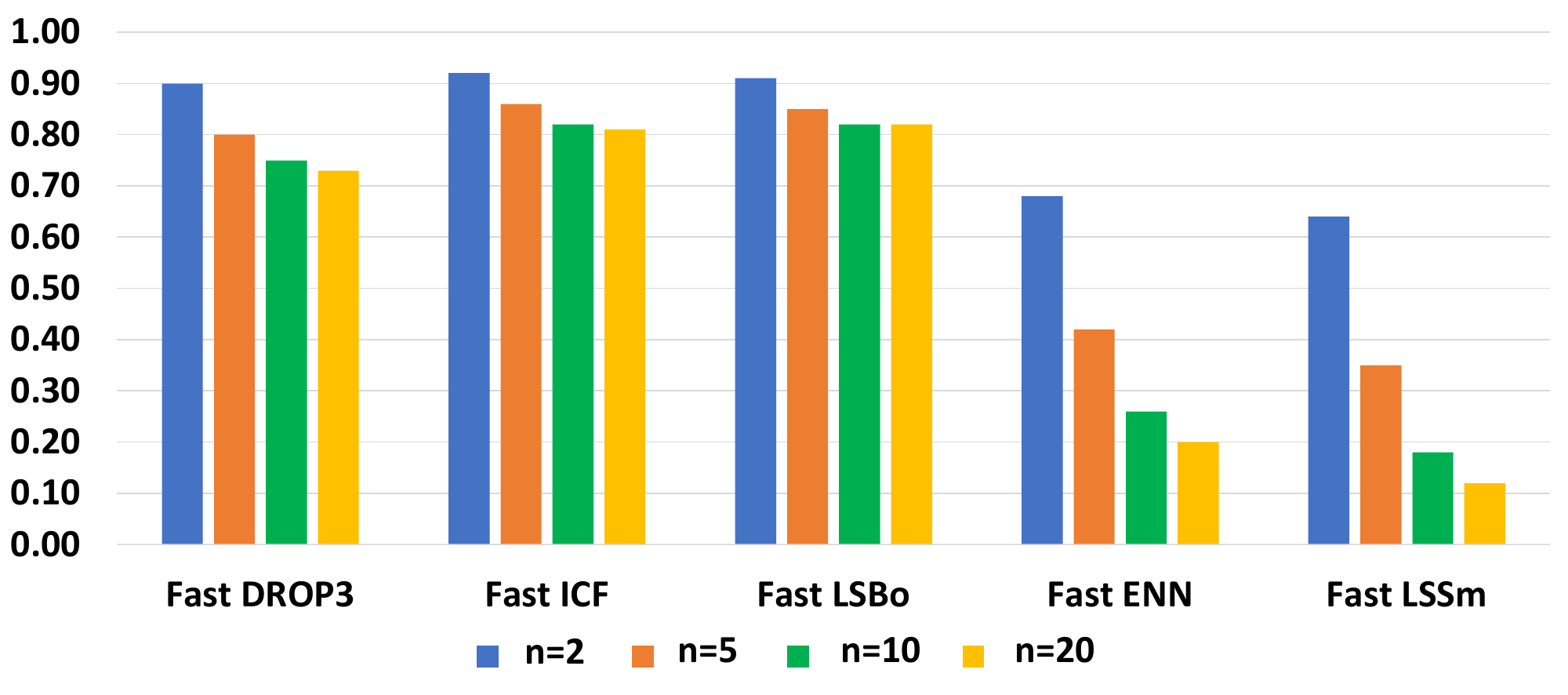}}
	\caption{Average reduction (considering the reduction in all datasets), for each value of $n$, of the 5 algorithms that use the PSASA algorithm as a pre-step.}
	\label{fig:reductionVaryingN}
\end{figure*}

We also evaluated the impact of the parameter $n$ on the performance (in terms of accuracy and reduction) of the PSASA algorithm. Figure \ref{fig:accuracyVaryingN} represents the \emph{average accuracy}, while Figure \ref{fig:reductionVaryingN} represents the \emph{average reduction} achieved by the proposed algorithms, as a function of the parameter $n$, with $n$ assuming the values 2, 5, 10 and 20. In this experiment, we also considered the 10-fold cross validation schema. The accuracy was measured considering the KNN classifier, with $k=3$. Notice that both figures, \ref{fig:accuracyVaryingN} and \ref{fig:reductionVaryingN}, present the \emph{average} performance (considering all the datasets) of each algorithm, with each value of $n$. 

The results represented in Figure \ref{fig:accuracyVaryingN} and \ref{fig:reductionVaryingN} suggest that as the value of $n$ increases, the accuracy also increases and the reduction decreases. This behavior is expected, since a higher value of $n$ allows the PSASA algorithm to create a higher quantity of finer-grained spatial partitions. Thus, as the value of $n$ increases, the number of prototypes created by the PSASA algorithm also increases, decreasing the level of \emph{spatial abstraction} achieved by the algorithm and increasing the amount of information that is preserved from the original dataset.

\begin{figure} [!ht]
	\centerline{\includegraphics[width=0.8\columnwidth]{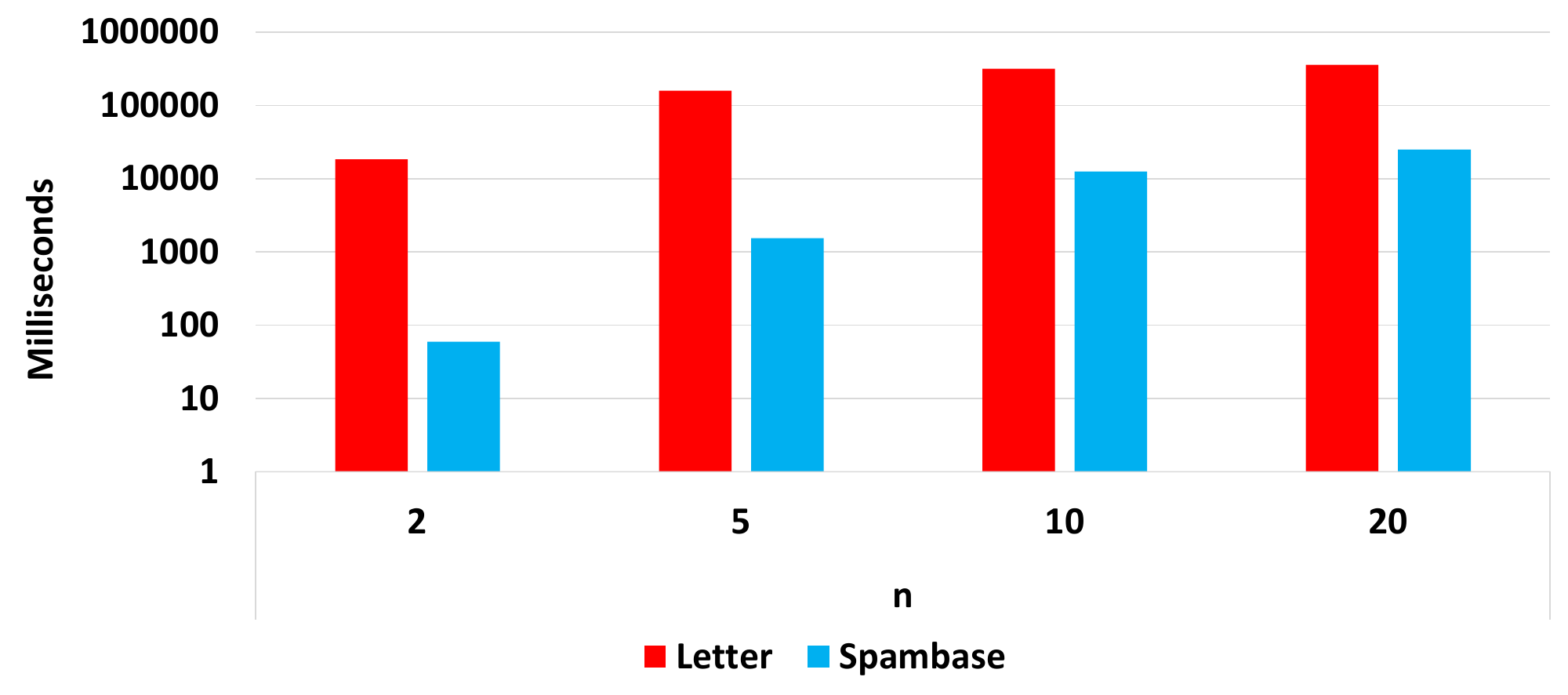}}
	\caption{Running times of the Fast DROP3 algorithm, according to different values of $n$.}
	\label{fig:timesDROP3}
\end{figure}

\begin{figure} [!ht]
	\centerline{\includegraphics[width=0.8\columnwidth]{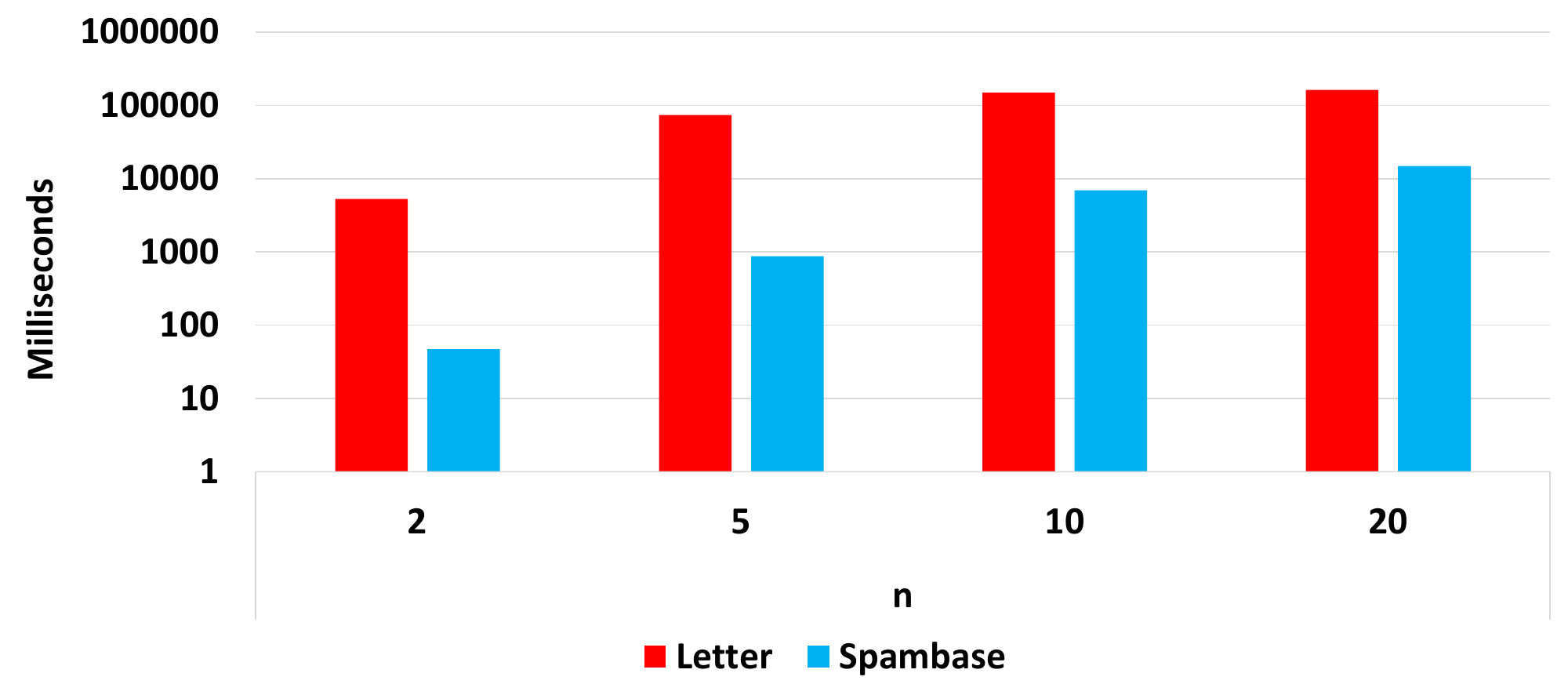}}
	\caption{Running times of the Fast ENN algorithm, according to different values of $n$.}
	\label{fig:timesENN}
\end{figure}

\begin{figure} [!ht]
	\centerline{\includegraphics[width=0.8\columnwidth]{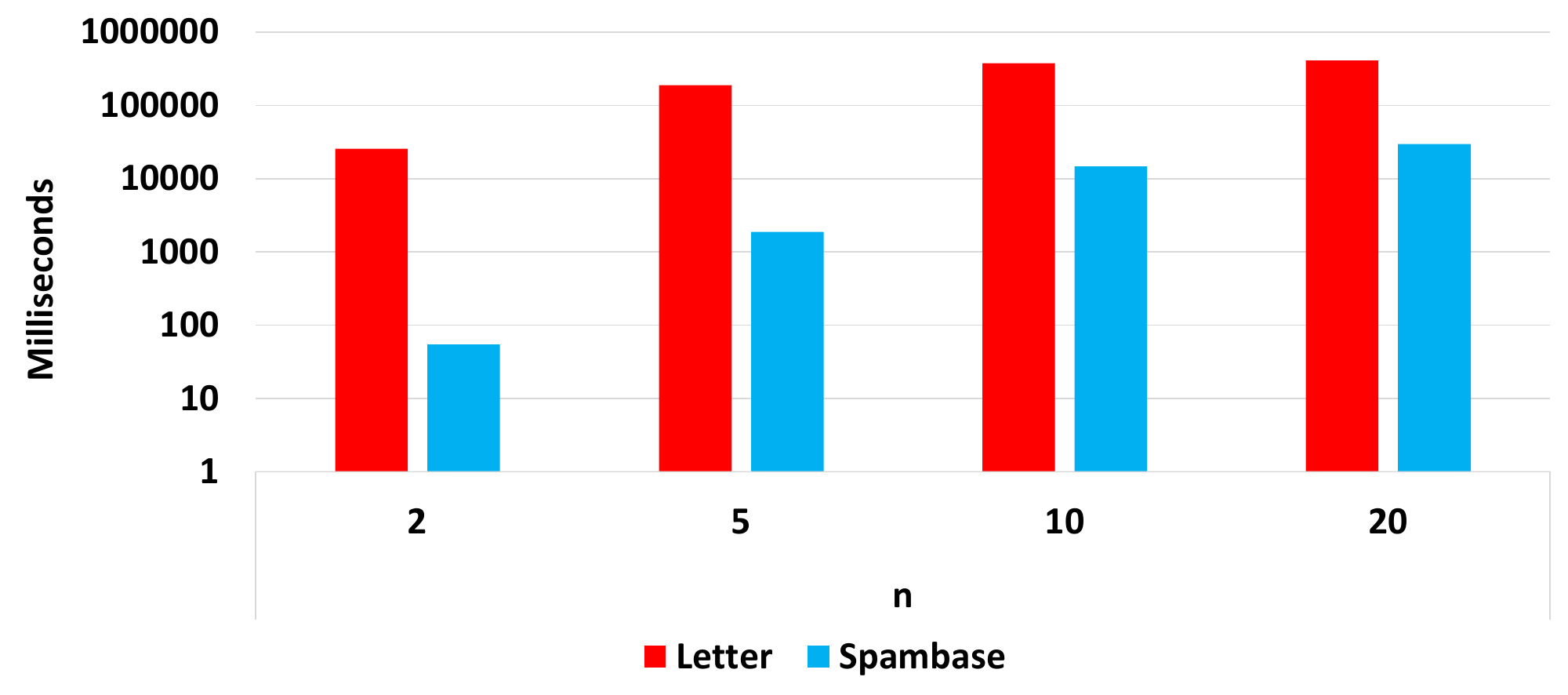}}
	\caption{Running times of the Fast ICF algorithm, according to different values of $n$.}
	\label{fig:timesICF}
\end{figure}

\begin{figure} [!ht]
	\centerline{\includegraphics[width=0.8\columnwidth]{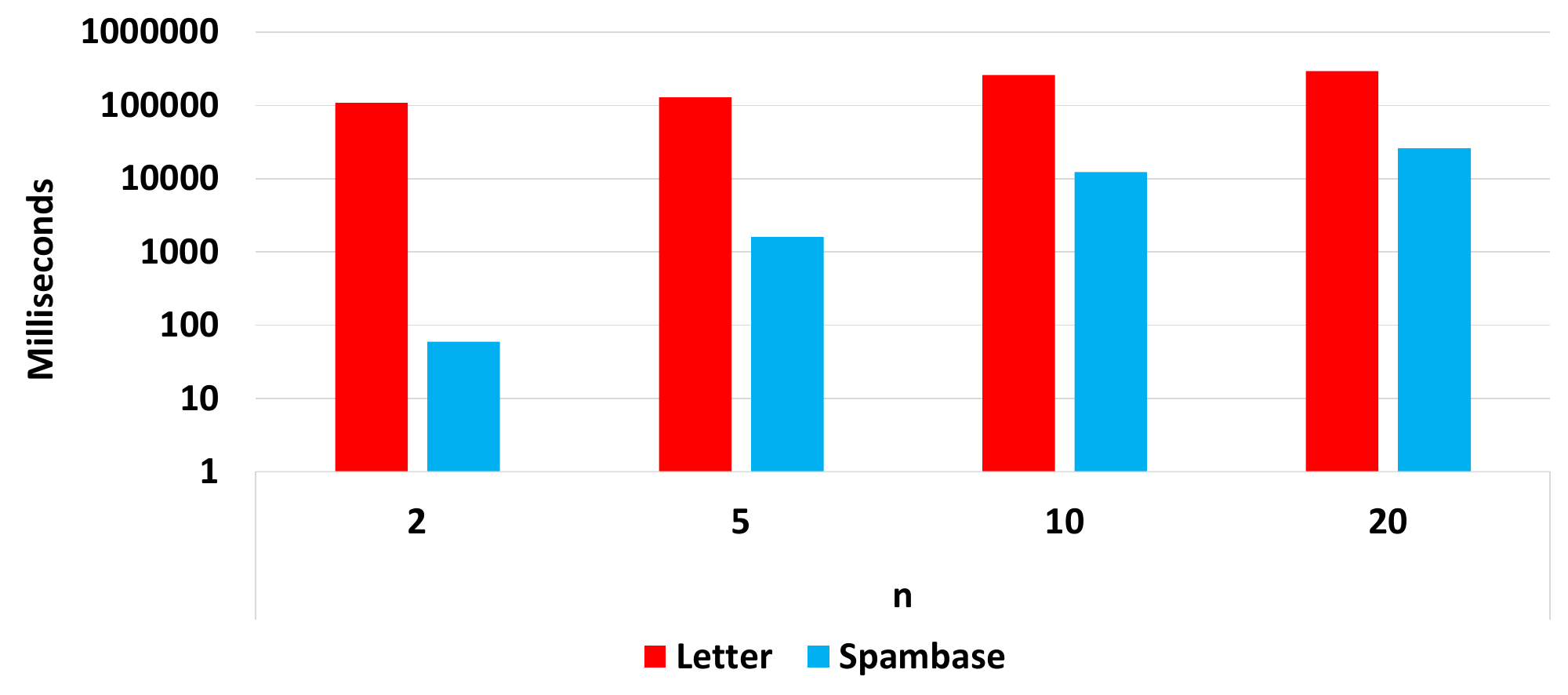}}
	\caption{Running times of the Fast LSBo algorithm, according to different values of $n$.}
	\label{fig:timesLSBO}
\end{figure}

\begin{figure} [!ht]
	\centerline{\includegraphics[width=0.8\columnwidth]{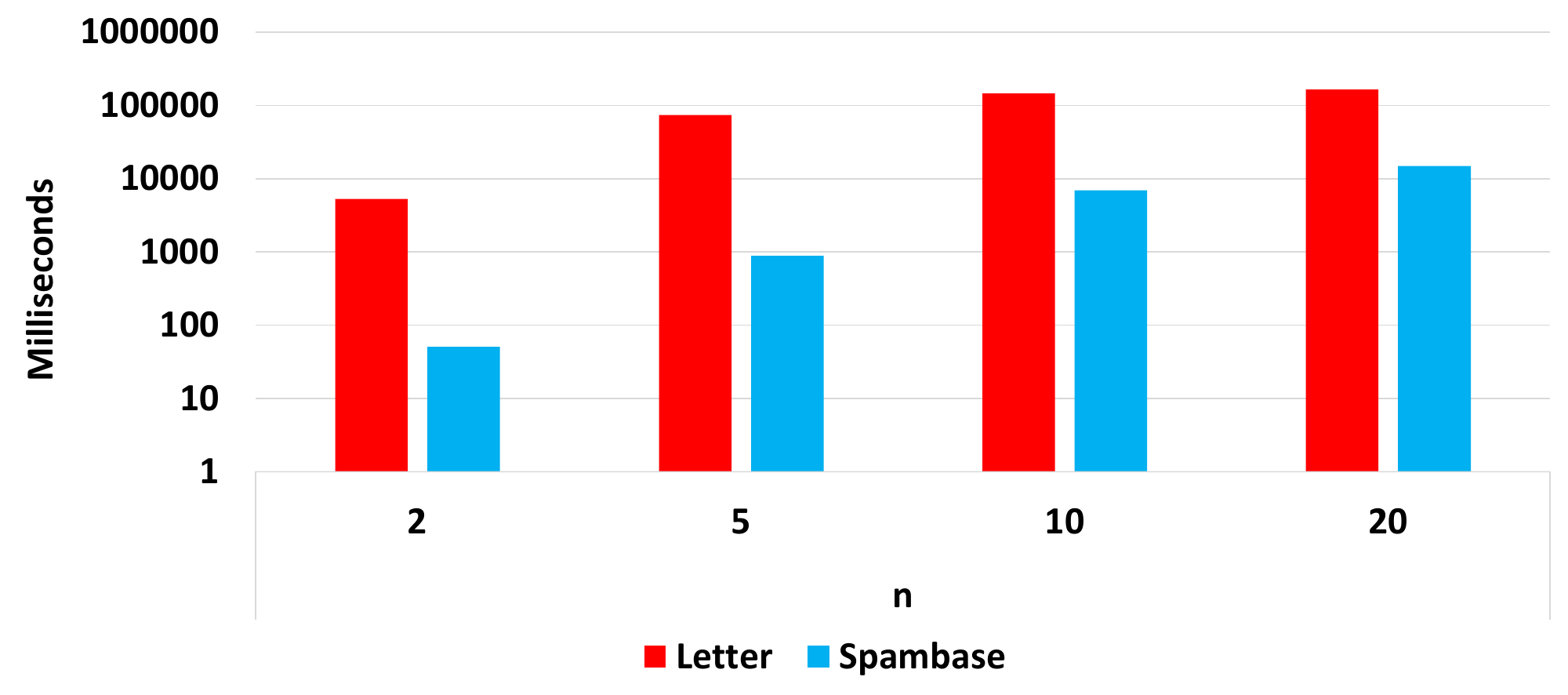}}
	\caption{Running times of the Fast LSSm algorithm, according to different values of $n$.}
	\label{fig:timesLSSM}
\end{figure}

Finally, we also carried out an experiment to evaluate the impact of the parameter $n$ on the running time of the algorithms that adopt the PSASA algorithm. This experiment was carried out in the same machine of the last experiment and adopted the same parametrization and the same datasets. We considered the following values of $n$: 2, 5, 10 and 20. The Figures \ref{fig:timesDROP3}, \ref{fig:timesENN}, \ref{fig:timesICF}, \ref{fig:timesLSBO} and \ref{fig:timesLSSM} represent the results of this experiment, for the Fast DROP3, Fast ENN, Fast ICF, Fast LSBo and Fast LSSm algorithm, respectively. Notice that the time axis adopts a \emph{logarithmic scale} (due to the big differences in performances) in these figures. According to the results of this experiment, we can observe that, in general, as the value of $n$ increases, the running times of the algorithms also increase. Notice that the same pattern can be observed in all the algorithms considered in the experiment. This behavior is expected, since as the value of $n$ increases, the number of spatial partitions identified by the PSASA algorithm also increases and, therefore, the number of prototypes identified by the PSASA also increases. And, as the number of prototypes produced by the PSASA algorithm increases, the running time of the prototype selection algorithm that is used in the final stage of the pipeline also increases, since their running times are proportional to the number of data points received as input. Thus, the value of $n$ affects the capability of the PSASA algorithm to reduce the search space that will be considered by the prototype selection algorithm that will be used in the subsequent step of the prototype selection pipeline.

In summary, the experiments show that the PSASA algorithm is able to speed up other conventional prototype selection algorithms, while preserving accuracy and reduction rates that are comparable (or better, in some cases) to those of the original algorithms (without adopting the PSASA algorithm). Besides that, the experiments show that the parameter $n$ affects the accuracy, the reduction rate, and the running time of the prototype selection pipelines that adopt the PSASA algorithm as a pre-processing step. As the value of $n$ increases, the accuracy increases, the reduction rate decreases, and the running time increases.
	
\section{Conclusion} \label{sec:conclusion}
	
In this paper, we proposed an efficient algorithm, called PSASA (Prototype selection accelerator based on spatial abstraction), which can be used for speeding up other conventional prototype selection algorithms. It adopts the notion of \emph{spatial partition}, which, in an overview, is a hyperrectangle within the space defined by the dataset. Firstly, the PSASA algorithm divides the dataset into a set of spatial partitions. After, it creates a set of prototypes for each spatial partition, where each prototype abstracts a set of instances that belong to that spatial partition. The algorithm takes as input the value $n$, which represents the number of intervals in which the dimensions of the dataset will be divided.   
	
Our experiments show that PSASA is able to reduce the running time of conventional prototype selection algorithms available in the literature, while preserving accuracy and reduction rates that are comparable to those of the original algorithms (without adopting the PSASA algorithm). These results suggest that PSASA is a promising solution for allowing conventional prototype selection algorithms to deal with big data and in contexts where a low running time is critical.
	
In future works, we plan to investigate how to allow the algorithm to automatically and efficiently identify the best way of dividing the dataset in a set of spatial partitions, without user intervention.



\bibliographystyle{unsrtnat}
\bibliography{references}  






\end{document}